\documentclass{article}
\usepackage{spconf,amsmath,graphicx}
\usepackage{caption}
\usepackage{subcaption}
\usepackage{algorithm}
\usepackage{algpseudocode}
\usepackage{amsmath}
\usepackage{bbm}
\usepackage{adjustbox}
\DeclareMathOperator*{\argmax}{arg\,max}

\algnewcommand{\Inputs}[1]{%
  \State \textbf{Inputs:}
  \Statex \hspace*{\algorithmicindent}\parbox[t]{.8\linewidth}{\raggedright #1}
}

\title{Task Selection and Assignment for Multi-modal Multi-task Dialogue Act Classification with Non-stationary Multi-armed Bandits}
\name{Xiangheng He$^{1}$, Junjie Chen $^{3}$, Björn W.\ Schuller$^{1,2}$}
\address{
  $^1$GLAM -- Group on Language, Audio, \& Music, Imperial College London, UK\\
  $^2$CHI -- Chair of Health Informatics, MRI, Technical University of Munich, Germany\\
  $^3$Department of Computer Science, The University of Tokyo, Japan}
\begin{document}
%
\maketitle
\begin{abstract}
Multi-task learning (MTL) aims to improve the performance of a primary task by jointly learning with related auxiliary tasks. Traditional MTL methods select tasks randomly during training. However, both previous studies and our results suggest that such a random selection of tasks may not be helpful, and can even be harmful to performance. Therefore, new strategies for task selection and assignment in MTL need to be explored. This paper studies the multi-modal, multi-task dialogue act classification task, and proposes a method for selecting and assigning tasks based on non-stationary multi-armed bandits (MAB) with discounted Thompson Sampling (TS) using Gaussian priors. Our experimental results show that in different training stages, different tasks have different utility. Our proposed method can effectively identify the task utility, actively avoid useless or harmful tasks, and realise the task assignment during training. Our proposed method is significantly superior in terms of UAR and F1 to the single-task and multi-task baselines with p-values \textless 0.05. Further analysis of experiments indicates that for the dataset with the data imbalance problem, our proposed method has significantly higher stability and can obtain consistent and decent performance for minority classes. Our proposed method is superior to the current state-of-the-art model.
\end{abstract}
\section{Introduction}
\label{sec:intro}


\begin{figure}[t]
    \centerline{\includegraphics[width=0.8\columnwidth]{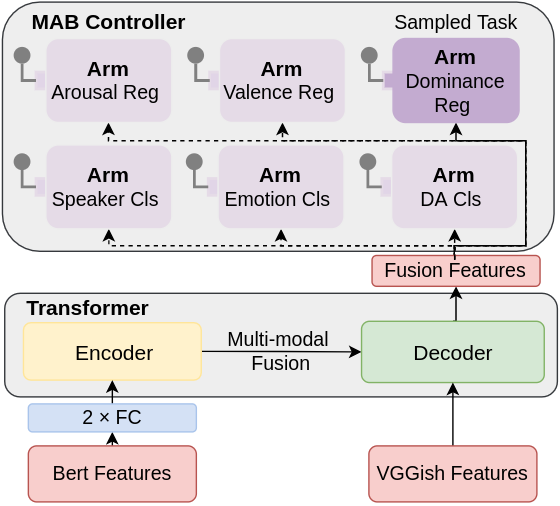}}
    \vspace{-0.2cm}
    \caption{
    Proposed model architecture.
    }
    \label{fig:model}
\end{figure}

Multi-task learning (MTL) \cite{DBLP:books/sp/98/Caruana98} aims to enhance the performance of a primary task by leveraging the knowledge of related auxiliary tasks. It has achieved success over many research areas \cite{zhang2019attention}\cite{liu2019end}. However, previous work \cite{liu2019loss}\cite{lee2016asymmetric} found that using auxiliary tasks without careful selection can have a negative impact on performance, a phenomenon referred to as negative transfer. Therefore, implementing effective MTL faces two challenges: task selection, defined as the need to identify truly useful tasks, and task assignment, defined as the need to decide how to allocate those tasks.

Dialogue Act Classification (DAC) is a task of determining the communicative intention a speaker is holding in an utterance. This task serves as the basis for many natural language understanding tasks \cite{higashinaka2006incorporating}\cite{cavazza2005dialogue} and is crucial in research domains such as dialogue systems \cite{higashinaka2014towards} and speech dialogue generation \cite{walker2001date}. Early work focused on single-task single-modal DAC \cite{khanpour2016dialogue}, that is, uses chat transcripts with only the text mode. Recently, thanks to the development of multi-modal open-source datasets \cite{saha2020towards}\cite{saha2021emotion}, multi-task multi-modal DAC has gained attention. However, existing methods \cite{saha2020towards}\cite{saha2021emotion} were based on the traditional MTL, which applied a random task selection strategy. Our results show that this strategy can produce negative transfer phenomena.

This paper focuses on the multi-task multi-modal DAC task. Our primary task is DAC (8-class) accompanied by five auxiliary tasks: emotion classification (5-class), speaker classification (10-class), arousal regression (range [1,5]), valence regression (range [1,5]), and dominance regression (range [1,5]). Our contributions are as follows: i. We propose a novel training strategy based on non-stationary MAB with discounted TS to deal with the above two challenges of task selection and task assignment; ii. We illustrate that the proposed method can effectively identify task utility in different training stages, avoid the negative transfer phenomena, and realise the end-to-end task assignment during training; iii. We demonstrate that the performance improvement mainly comes from the significantly high stability of the minority classes; iv. We compare the performance of the proposed method with the current state-of-the-art method \cite{saha2021emotion} (SOTA) and show the superiority of the proposed method.

\section{Background and Related Works}
\label{sec:related}

\subsection{MAB with TS}
\label{ssec:MAB_with_TS}

The above-mentioned two challenges of MTL are potentially solved by MAB due to its sequential decision-making properties. Assume that the MAB problem has $K$ arms $\mathcal{I}:= \{1, 2, ..., K\}$ with a limited time horizon $T$. At each round $t$, the agent has to select an arm to play. Each arm when played will observe a reward $r_t$ according to an unknown reward distribution with support in $[0, 1]$. The objective is to maximise the cumulative reward $\sum_{t=1}^{T} r_t$ over $T$ periods. For each arm, TS assumes a prior distribution on the expected reward of the unknown reward distribution, and plays the arm with the highest value of the expected reward at each round via sampling. There are two common versions of TS, namely Beta priors with Bernoulli likelihood \cite{DBLP:conf/naacl/GuoPB19} and Gaussian prior with Gaussian likelihood \cite{DBLP:conf/aistats/AgrawalG13}. Depending on whether the unknown reward distribution of each arm changes over time, there are non-stationary MAB \cite{DBLP:journals/corr/abs-2305-10718} and stationary MAB \cite{DBLP:conf/aistats/AgrawalG13}.

\subsection{Related Works}
\label{ssec:related_works}
Previous work for multi-task DAC \cite{saha2020towards}\cite{saha2021emotion} showed that the introduced emotion recognition (ER) task achieved better performance than their single-task baseline. However, we obtained the opposite result. Aiming to further reveal the impact of emotional information on DAC performance, we introduce four emotion-related tasks which consist of the ER task and three other emotion regression tasks. \cite{saha2020towards}\cite{saha2021emotion} applied a uniformly random task selection strategy during training, whereas we use a novel MAB-based task selection and assignment strategy.

\cite{DBLP:conf/naacl/GuoPB19} first proposed a paradigm of task selection for MTL via MAB with TS. However, it continuously added strong bias to the primary task, thereby failing to explore the different utility of auxiliary tasks in different training stages. It only exploited MAB for task selection and then needed to combine it with a Gaussian Process to realise task assignment. Our method can identify the utility of tasks in different training stages and can achieve the end-to-end task assignment.


\section{Method}
\label{sec:method}

\begin{algorithm}[t]
\caption{Non-stationary MAB with discounted TS}\label{alg:MAB}
\begin{algorithmic}
\Inputs{$\gamma$, $\hat{\mu}_1(i)$, $\tilde{\mu}_1(i)$, $N_1(i)$, s, $\tau_1(i)$, $\tau_{max\_bound}$}
\For{$t = 1, ..., T$} 
    \For{$i = 1, ..., K$} 
    \State sample $\theta_t(i)$ from $\mathcal{N}(\hat{\mu}_t(i), \tau_t(i)^2)$
    \EndFor
    \State Select task $i_t=\argmax_i \theta_t(i)$ and get reward $r_t(i_t)$
    \For{$i = 1, ..., K$}
    \State $\tilde{\mu}_{t+1}(i)=\gamma\tilde{\mu}_{t}(i)+\mathbbm{1}_{i=i_t}(i)r_t(i_t)$
    \State $N_{t+1}(i)=\gamma N_{t}(i)+\mathbbm{1}_{i=i_t}(i)$
    \State $\hat{\mu}_{t+1}(i)=\frac{\tilde{\mu}_{t+1}(i)}{N_{t+1}(i)}$
    \State $\tau_{max} = \min\{st+\tau_1(i),\tau_{max\_bound}\}$
    \State $\tau_{t+1}(i)=\min\{\frac{1}{\sqrt{N_{t+1}(i)}},  \tau_{max}\}$
    \EndFor
\EndFor
\end{algorithmic}
\end{algorithm}

Inspired by \cite{DBLP:conf/naacl/GuoPB19}, we model the MTL as a MAB problem with TS. $K$ arms correspond to $K$ tasks. Selecting an arm to play corresponds to selecting a task to train the model. We define the utility of the selected task as the performance improvement after selecting this task to update the model relative to the performance of the current best model. That is, reward $r_t = V_t(i_t) - V^{best}_t$, where $V_t(i_t)$ refers to the validation performance of the model obtained after training with the selected task $i_t$ at the round $t$ and $V^{best}_t$ refers to the best performance obtained on the validation set before round $t$. So the larger the reward, the greater the performance improvement (utility) brought by the selected task $i_t$. The intrinsic utility of each task may change during training, e.\,g., the primary task might be more useful at the beginning, but not necessarily at the end. Therefore, the agent faces a non-stationary system where it must consistently be motivated to explore, ensuring it can adapt to shifts as the system evolves. We make small changes to the algorithm in \cite{DBLP:journals/corr/abs-2305-10718}. Algorithm \ref{alg:MAB} shows the details, where $\gamma$ denotes discounted factor, $\tau_t(i)$ denotes the sampling variance, $\tau_{max\_bound}$ denotes the upper bound of the sampling variance, $\hat{\mu}_t(i)$ denotes the estimation of the expected reward of task $i$ at round $t$, $\tilde{\mu}_1(i)$ denotes the discounted cumulative reward, $N_t(i)$ denotes the discounted number of plays of task $i$ until round $t$, and $s$ denotes the slope of $\tau_{max}$. The key to the algorithm is to encourage exploration by increasing the sampling variance of unselected tasks and decreasing the sampling variance of selected tasks.

Figure \ref{fig:model} shows the proposed model architecture. We use the encoder and decoder of the standard transformer to receive the features of two modalities and perform self-attention fusion. The MAB controller part selects a task to be trained.


\section{Experimental Setup}
\label{sec:setup}

\begin{figure}[t]
    \centerline{\includegraphics[width=8cm]{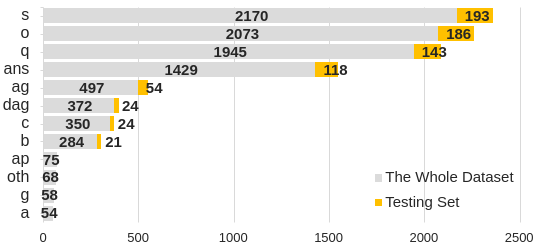}}
    \vspace{-0.1cm}
    \caption{
    Class distribution for the whole dataset and the testing set we use.
    }
    \label{fig:data_distribution}
\end{figure}

\textbf{Dataset: }We use the database proposed in \cite{saha2021emotion} for all our experiments. This dataset is an extension of the original IEMOCAP dataset \cite{DBLP:journals/lre/BussoBLKMKCLN08}, adding 12 dialogue act (DA) tags including Greeting (g), Apology (ap), Command (c), Question (q), Answer (ans), Agreement (ag), Disagreement (dag), Statement-Opinion (o), Statement-Non-Opinion (s), Acknowledge (a), Backchannel (b), and Others (oth). The dataset has a total of 9376 audio samples. Figure \ref{fig:data_distribution} shows its class distribution.

The dataset used in Section \ref{ssec:task} to Section \ref{ssec:stability} has 8 DA classes. We removed four DA classes
(`ap', `oth', `g', `a') with less than 0.8\% of the total number of data samples. For emotion classification task, we replace the `excitement' class in IEMOCAP with `happiness' as in \cite{xia2015multi}, and only consider classes `anger', `sadness', `neutral', `frustration', and `happiness'. The final dataset has 7623 data samples in total. We did a random split with 70\%, 20\%, and 10\% for training, validation, and testing sets, respectively. 
The distribution of 8 DA classes in the testing set is shown in Figure \ref{fig:data_distribution}. We call the 4 classes (`s', `o', `q', `ans') with more than 117 data samples the majority class, and the 4 classes (`ag', `dag', `c', `b') with less than 55 data samples the minority class. The dataset utilised in Section \ref{sssec:SOTA} has all 12 DA classes. To achieve a fair comparison, we utilise the same dataset and split as in \cite{saha2021emotion}, allocating 80\% for training and 20\% for testing.


\noindent \textbf{Experimental Details: }The baseline methods we implement include a single-task baseline (ST) and a multi-task baseline (MT). ST only uses the primary task during training. MT randomly selects a task from all 6 tasks in each epoch of training to update the model. Our proposed MAB method selects a task from 6 tasks according to the MAB controller in each epoch of training to update the model.


We apply the pre-trained BERT model \footnote{https://huggingface.co/bert-base-cased} to extract textual features and apply the pre-trained vggish model \footnote{https://github.com/tensorflow/models/tree/master/research/audioset/vggish} to extract audio features. We fix the sequence length of text features to 60 and the length of audio features to 20 by truncation or padding. The transformer part has 8 heads with 6 encoder and decoder layers. We choose cross-entropy loss for classification tasks and CCC loss for regression tasks. We apply the Adam optimiser with a weight decay of 0.001. We train each model with 500 epochs and choose the model with the best validation loss. As in \cite{DBLP:conf/naacl/GuoPB19}, we initially set a slightly stronger prior for the primary task $i^p$, which $\hat{\mu}_1(i^p)$ and $\tilde{\mu}_1(i^p)$ are set to 0.3. $\hat{\mu}_1(i^a)$ and $\tilde{\mu}_1(i^a)$ for auxiliary tasks $i^a$ are set to 0. $N_1(i)$, $\tau_1(i)$, $\tau_{max\_bound}$, and $\gamma$ for all tasks are set to 0, 0.05, 0.5, and 0.9, respectively.

\begin{figure}[t]
    \centerline{\includegraphics[width=9cm]{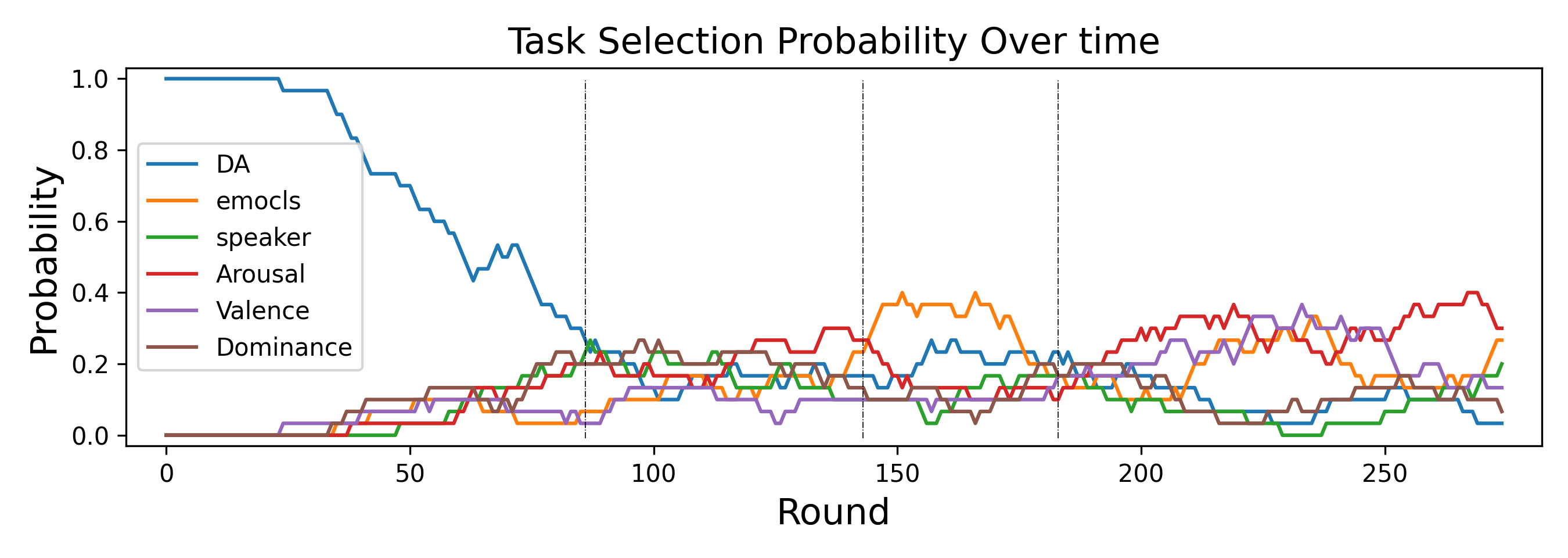}}
    \vspace{-0.45cm}
    \caption{
    Task selection during the training process for the proposed MAB method.
    }
    \label{fig:task_selection}
\end{figure}

\begin{table}[]
\begin{center}
\caption{Overall performance (8-class) on the testing set. Mean performance and standard deviations of 10 trials are reported.}
\begin{tabular}{l|ccc}
\hline
 & UAR                  & F1                   & Accuracy \\
\hline
ST        & 0.618±0.061          & 0.604±0.070          & 0.675±0.022          \\ \hline
MT          & 0.522±0.052          & 0.499±0.054          & 0.608±0.019          \\ \hline
Our            & \textbf{0.667±0.013} & \textbf{0.655±0.013} & \textbf{0.683±0.010} \\ \hline
\end{tabular}
\label{tab:overall_performance}
\end{center}
\end{table}

\section{Results and Analysis}
\label{sec:results}

\subsection{Task Selection}
\label{ssec:task}


Figure \ref{fig:task_selection} shows the probability of each task selected during training. We compute the probability for task $i$ using the number of times that task $i$ is selected in a sliding window of length 30. As shown by the gray line in the figure, the training process can be divided into 4 stages. In the first stage, the model first tends to choose the primary task that is most likely to bring the largest reward. As time goes on, the model finds that the reward of choosing only the primary task starts to decline, so it begins to explore other tasks and starts a comprehensive exploration in the second stage. In the third stage, the model finds that the emotion classification task is the most useful task, so there is a greater probability of selecting the emotion classification task. Finally, in the fourth stage, the model considers the arousal regression and the valence regression tasks as the first two useful tasks. The speaker classification and dominance regression tasks have a low selection probability during the entire training process, indicating that these two tasks are of low utility to the primary task.

\subsection{Overall and Sub-class Performance}
\label{ssec:overall}

\textbf{Overall Performance: }Table \ref{tab:overall_performance} shows the overall performance. The proposed method outperforms the two baseline methods on all metrics. The performance of the MT is far worse than that of the ST, which indicates that the random selection strategy can impede performance. The proposed method has a slight improvement in accuracy (metric not considering the class distribution) but has a significant improvement in F1 \footnote{The F1 scores we mentioned in the paper are all Macro-F1 scores.} and UAR (metrics considering the class distribution) when compared to the two baselines. Our significance experiment (Welch's t-test \cite{welch1947generalization} at 5\,\% significance level) shows that this improvement of F1 and UAR is statistically significant. To find out the reason, we perform an analysis of sub-class performance.


\begin{figure}[t]
    \vspace{-0.2cm}
     \centering
     \begin{subfigure}[b]{0.9\columnwidth}
         \centering
         \includegraphics[width=\columnwidth]{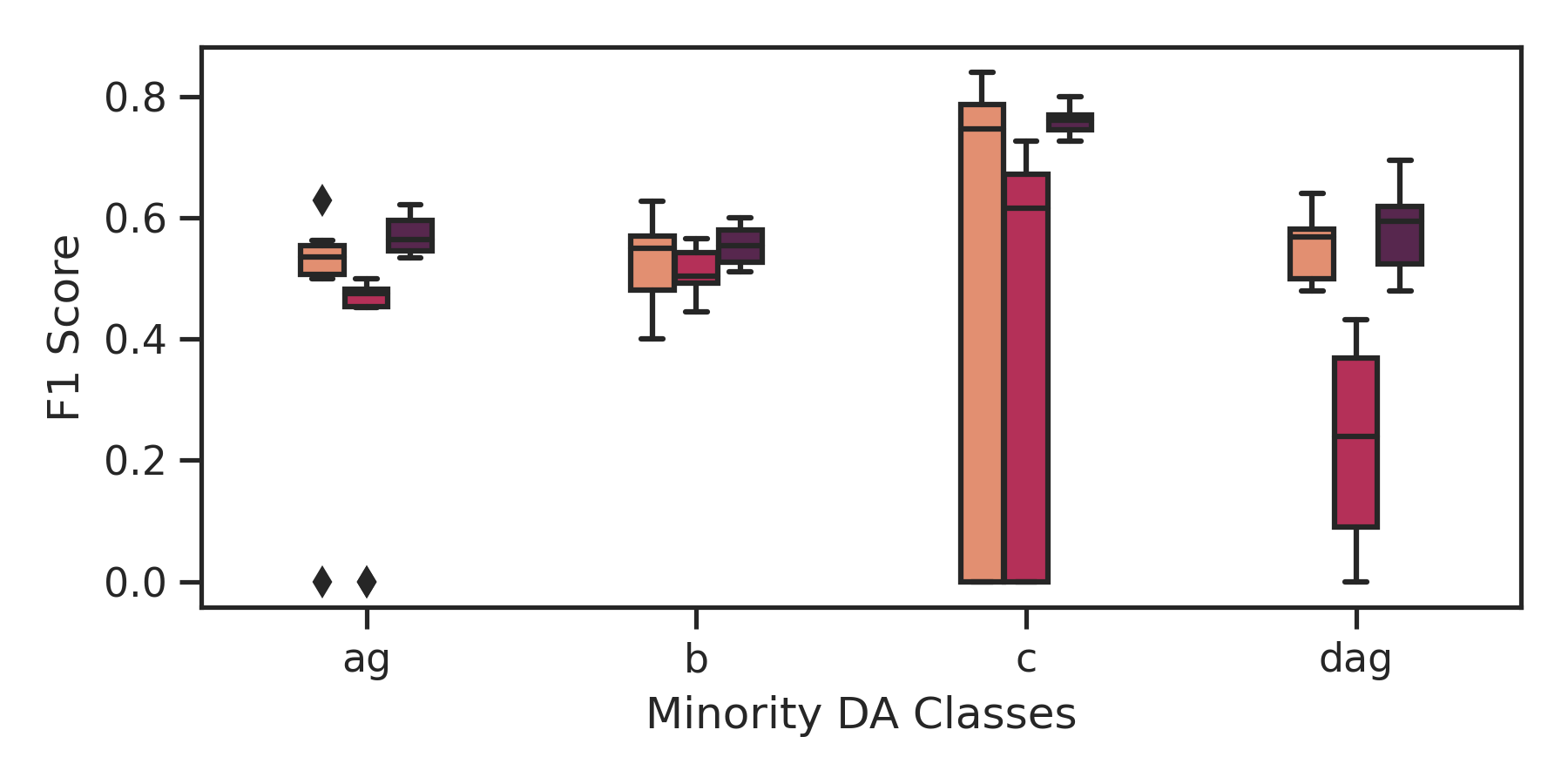}
         \vspace{-0.7cm}
         \label{fig:minority_class_F1}
     \end{subfigure}
     \hfill
     \begin{subfigure}[b]{\columnwidth}
         \centering
         \includegraphics[width=0.9\columnwidth]{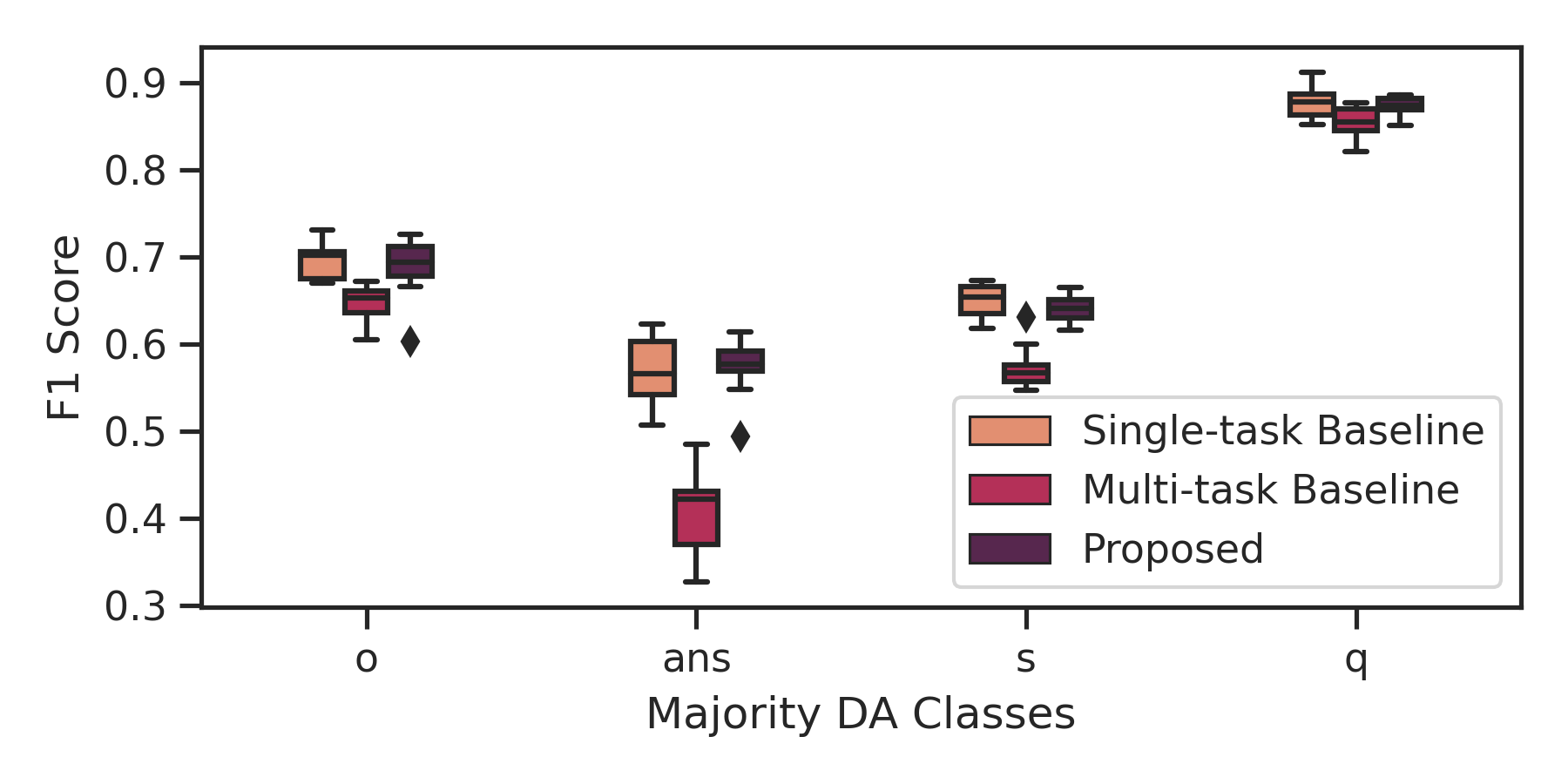}
         \vspace{-0.4cm}
         \label{fig:majority_class_F1}
     \end{subfigure}
        \caption{Boxplot of sub-class F1 on the testing set for 10 trials.}
        \label{fig:per_class_F1}
\end{figure}


\begin{table}[t]

\begin{center}
\caption{F1 for 4 minority classes on the testing set. Mean performance and standard deviations of 10 trials are reported.}
\begin{adjustbox}{width=\columnwidth}
\begin{tabular}{l@{\hskip 0.05cm}|@{\hskip 0.05cm}c@{\hskip 0.05cm}c@{\hskip 0.05cm}c@{\hskip 0.05cm}c}
\hline
            & ag                   & c                    & dag                  & b                    \\
\hline
ST & \begin{tabular}[c]{@{}l@{}}\phantom{a}0.489±0.176\end{tabular}          & \begin{tabular}[c]{@{}l@{}}\phantom{a}0.473±0.408\end{tabular}          & \begin{tabular}[c]{@{}l@{}}\phantom{a}0.550\textbf{±0.054}\end{tabular}          & \begin{tabular}[c]{@{}l@{}}\phantom{a}0.530±0.068\end{tabular}          \\ \hline
MT  & \begin{tabular}[c]{@{}l@{}}\phantom{a}0.382±0.202\end{tabular}          & \begin{tabular}[c]{@{}l@{}}\phantom{a}0.398±0.344\end{tabular}          & \begin{tabular}[c]{@{}l@{}}\phantom{a}0.222±0.163\end{tabular}          & \begin{tabular}[c]{@{}l@{}}\phantom{a}0.510±0.040\end{tabular}          \\ \hline
Our    & \textbf{\begin{tabular}[c]{@{}l@{}}\phantom{a}0.570±0.031\end{tabular}} & \textbf{\begin{tabular}[c]{@{}l@{}}\phantom{a}0.759±0.021\end{tabular}} & \begin{tabular}[c]{@{}l@{}}\phantom{a}\textbf{0.578}±0.068\end{tabular} & \textbf{\begin{tabular}[c]{@{}l@{}}\phantom{a}0.554±0.031\end{tabular}} \\
\hline
\end{tabular}
\end{adjustbox}
\label{tab:minority_class_F1}
\end{center}
\end{table}

\begin{table}[t]
\begin{center}
\caption{
Overall Performance (12-class) for the current SOTA and our proposed MAB method.
}
\begin{tabular}{l|cc}
\hline
                  & Accuracy & F1     \\ \hline
SOTA early fusion \cite{saha2021emotion} & 0.6601  & 0.6385 \\ \hline
SOTA late fusion \cite{saha2021emotion} & \textbf{0.6963}  & 0.6786 \\ \hline
Our          & 0.6951  & \textbf{0.6858} \\ \hline
\end{tabular}
\label{tab:SOTA_compare}
\end{center}
\end{table}

\noindent \textbf{Sub-class Performance: }Figure \ref{fig:per_class_F1} shows the box plot of the sub-class F1 scores. The F1 score of the MT method is significantly worse than the other two methods in almost all classes. The performance of the proposed method is not significantly different from that of the ST method in four majority classes (the bottom part of Figure \ref{fig:per_class_F1}). However, for minority classes (the upper part of Figure \ref{fig:per_class_F1}), the proposed method can obtain better results. Table \ref{tab:minority_class_F1} shows their sub-class F1 scores. For classes `dag' and `b', the proposed method has slightly better F1 scores than the ST method, while for classes `ag' and `c', the proposed method has much better results.

\subsection{Stability of Proposed Method}
\label{ssec:stability}

We can observe an interesting phenomenon in class `c'. Both Figure \ref{fig:per_class_F1} and Table \ref{tab:minority_class_F1} demonstrate a huge instability of the ST method and MT method in the prediction of class `c'. In fact, among the 10 ST models and 10 MT models recorded, 4 models each have an F1 score of 0. On the contrary, the proposed method has consistent and decent predictions.

The proposed method also has better stability for other classes besides class `c'. Figure \ref{fig:per_class_F1} and the standard deviation in Table \ref{tab:minority_class_F1} shows that the proposed method exhibits significantly improved stability in 3 out of 4 minority classes (`ag', `c', `b') compared to the ST, surpasses the stability of the MT across all 4 minority classes, and surpasses the two baseline methods in 3 out of 4 majority classes (`ans', `s', `q').
Table \ref{tab:overall_performance} also shows that the two baseline methods have significantly larger standard deviations in F1 and UAR, indicating that the overall stability of the proposed method is better than that of the two baseline methods.




\subsection{Comparison to SOTA}
\label{sssec:SOTA}

As shown in Table \ref{tab:SOTA_compare}, although the proposed method obtains a matching accuracy with the SOTA \cite{saha2021emotion}, it has a considerable improvement on the F1 score. Note that the 12-class dataset used here has a more severe data imbalance problem, which shows the superiority of our proposed method.

\section{Conclusion}
\label{sec:conclusion}
This paper studied the multi-task multi-modal DAC task and proposed a novel training strategy based on non-stationary MAB with discounted TS to deal with the task selection and task assignment challenges in MTL. Our results demonstrated that the proposed method can effectively identify task utility in different training stages and realise the end-to-end task assignment during training. The proposed method exhibited significant improvement in terms of UAR and F1 to both single-task and multi-task baselines. Experimental analysis of sub-class performance indicated that this improvement stemmed from the more stable performance of the minority classes. Further analysis illustrated that the overall stability of the proposed method is better than the baselines. The proposed method outperformed the current SOTA. Future work should also consider verifying the generalisability of the method, such as investigating primary tasks from diverse research fields.

\vfill\pagebreak


\bibliographystyle{IEEEbib}
\bibliography{strings,refs}

\end{document}